\documentclass[conference]{IEEEtran}
\usepackage[utf8]{inputenc}
\usepackage{graphicx}
\usepackage{hyperref}
\usepackage{makecell}
\graphicspath{ {images} }
\usepackage{float}
\usepackage{array}
\def\BibTeX{{\rm B\kern-.05em{\sc i\kern-.025em b}\kern-.08em
    T\kern-.1667em\lower.7ex\hbox{E}\kern-.125emX}}
\begin{document}

\title{Estimating Task Completion Times for Network Rollouts using Statistical Models within Partitioning-based Regression Methods\vspace{-0.2em}}
\author{
 \IEEEauthorblockN{Venkatachalam Natchiappan\IEEEauthorrefmark{1}, 
    Shrihari Vasudevan\IEEEauthorrefmark{1} and Thalanayar Muthukumar \IEEEauthorrefmark{2}}
\IEEEauthorblockA{
\textit{Ericsson}, \IEEEauthorrefmark{1}\textit{India} and \IEEEauthorrefmark{2}\textit{USA}\\
$\{n.venkatachalam \vert shrihari.vasudevan \vert thalanayar.muthukumar\}@ericsson.com$}}

\maketitle

\vspace{-20em}
\begin{abstract}
This paper proposes a data and Machine Learning-based forecasting solution for the Telecommunications network-rollout planning problem. Milestone completion-time estimation is crucial to network-rollout planning; accurate estimates enable better crew utilisation and optimised cost of materials and logistics. Using historical data of milestone completion times, a model needs to incorporate domain knowledge, handle noise and yet be interpretable to project managers. This paper proposes partition-based regression models that incorporate data-driven statistical models within each partition, as a solution to the problem. Benchmarking experiments demonstrate that the proposed approach obtains competitive to better performance, at a small fraction of the model complexity of the best alternative approach based on Gradient Boosting. Experiments also demonstrate that the proposed approach is effective for both short and long-range forecasts. The proposed idea is applicable in any context requiring time-series regression with noisy and attributed data.
\end{abstract}

\begin{IEEEkeywords}
Network-rollout planning, Forecasting, Tree/partition based regression
\end{IEEEkeywords}

\section{Introduction}
Mobile telecommunications technology is continuously evolving. The quest for supporting greater multimedia content at increased speed and scale (more consumers and devices) has resulted in significant advancements in both hardware (e.g., radio base stations, antennas and related technologies) and software (e.g., signal processing, effective use of different spectrum bands, service prioritization, etc.). These advancements and increased consumption of telecommunications services fuel greenfield (i.e., new) and brownfield (e.g., hardware upgrade) deployments requiring network-rollout activities. This paper focuses on developing predictive models that would enable effective planning for network rollouts.

Communications service providers (CSPs) provide Telecommunications services (e.g., messaging, telephony) to consumers; vendors build or manage the networks for the CSPs. Network rollouts occur in stages or phases, with each phase involving multiple milestones. Exemplar phases in typical network-rollouts include solution design, site acquisition, site engineering, civil works, installation, commissioning, configuration, integration, and acceptance\footnote{\href{https://www.youtube.com/watch?v=DQUbgLxqlXw}{Ericsson Network Roll-Out In a Nutshell}}.
In any project, a delay in one of the milestones will impact the completion time of every milestone that follows it. The time duration between two milestones in a project is sensitive to multiple external factors, not all of which are controllable by project managers. A few examples of such factors are weather conditions, unforeseen unavailability of site-engineers or equipment and delays due to regulatory controls for unpredictable situations such as birds nesting on the tower. These factors need to be accounted for while allocating people and equipment for a network rollout. A tool that enables project managers to optimally plan resource utilization can enable significant cost savings for the Telecommunications industry.

Each network rollout project contains a certain set of phases based on the nature of the project; these may involve radio access network expansion, rollout of optical-fiber networks, upgrading networks - e.g., 3G to 4G, and the optimisation of network parameters to fix quality issues. The phases and consequent milestones within any project are dependent on the specific contract between a CSP and a vendor. Thus, every project at every site has attributes characterizing its metadata.

Traditionally, project managers have target dates (dictated by the contract) for the completion of each milestone. Project managers track these using an Excel spreadsheet, for a future time-period. Due to a high volume of deployment projects simultaneously executed by vendors, project managers are only able to track milestone completions occurring in a limited future period. Hence, this manual approach is highly tedious and affords only limited forward planning capabilities. This paper proposes a data and Machine Learning (ML) solution, that provides accurate short and long-range forecasts, to enable effective project planning.

Data-sets in such contexts are thus composed of attributed time-series data. Several effective first-line approaches exist for time-series forecasting; for instance, ARIMA, SARIMA, Holt-Winters \cite{hyndman2018forecasting} etc. are time-tested methods. ARIMA and its variants cannot natively handle attributed time-series data. For regression problems (including forecasting) that involve attributed data, Gradient Boosting (GB) variants (e.g., XGBoost \cite{chen2016xgboost}) and Random Forests (RF) are well-established approaches. The former are particularly effective due to power of additive modeling. Most classical time-series forecasting methods are statistical approaches, that make use of the statistical (e.g., linear) relationship between predictors and target variable whereas GB, RF and other such partition-based approaches are piece-wise constant in the leaves of different stubs (base estimators).
Our proposed approach combines the best from both these approaches. Tree-based approaches are used to effectively partition attributed time-series data and at the leaves, time-series regression models are employed to leverage patterns in the data. Standard ML libraries (e.g. Scikit-Learn \footnote{https://scikit-learn.org/}) do not offer statistically-enhanced data-partitioned regression algorithms currently.

The contribution of this paper is to develop a forecasting solution that would enable a Telecommunications network rollout planning application. The proposed approach incorporates data trends and domain expectations to build competitive to better models, at marginal model complexity, relative to the best known approaches for the data. The approach is generic and possess significant application potential in any application context that involves attributed time-series data.

\begin{figure}[h]
\centering
\includegraphics[width=0.45\textwidth]{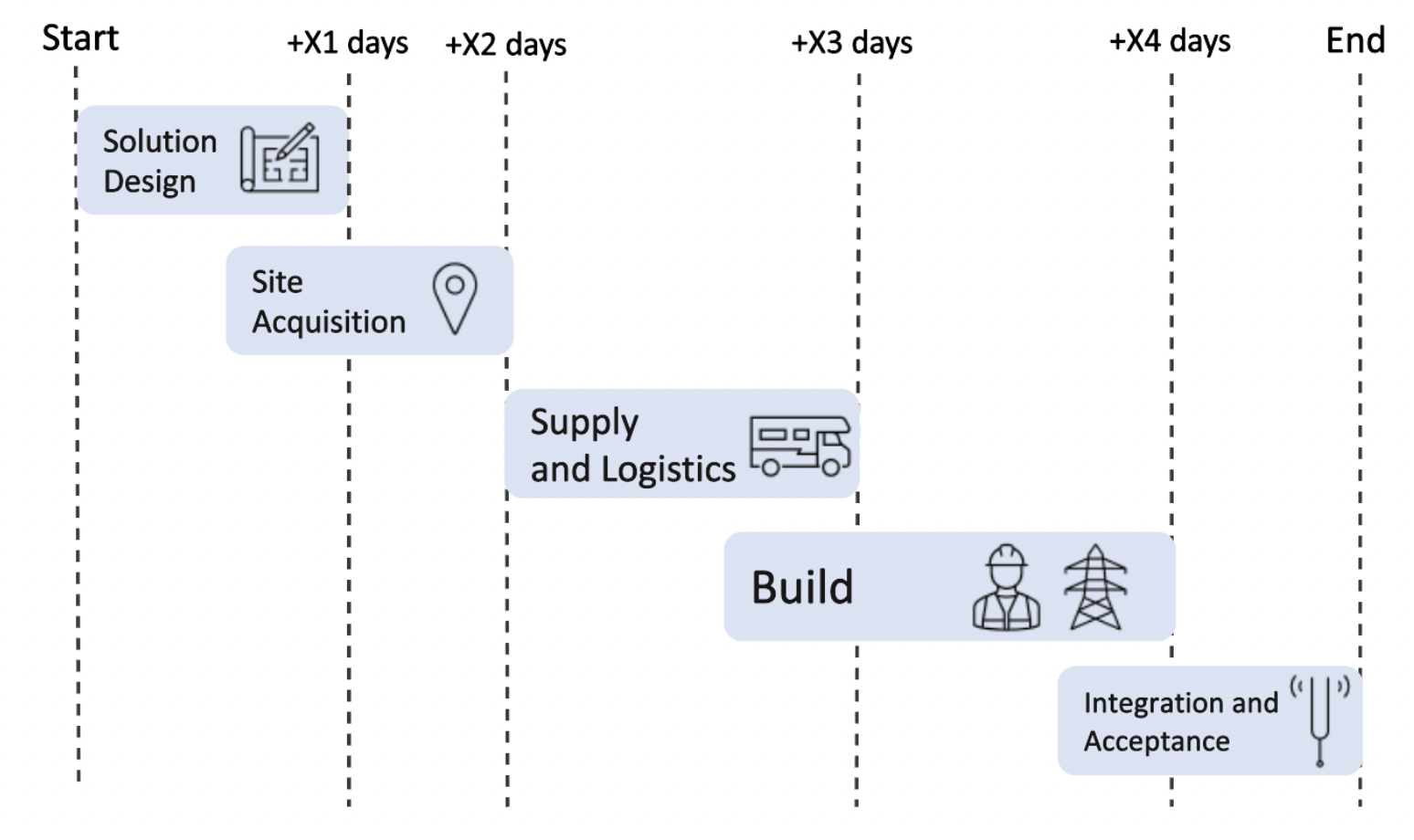}
\caption{Example Gantt chart for a network rollout project, depicting the phases' timelines. The typical overlap between the phases is shown. The Build phase is a superset for activities like site engineering, civil works and installation phases}
\label{fig1}
\end{figure}

\section{Problem definition and challenges}
\label{sec_prob_defn_n_challenges}

The problem addressed in this paper is that of estimating the time to complete a target milestone from a source milestone, given the site’s metadata, project’s metadata, and completion time of previously completed milestones, also referred to here as intermediate milestones. Each milestone in a phase may have a dependency on a few other milestones across different phases; hence, it may be performed only after the successful completion of all dependent milestones.
Certain milestones within a phase occur simultaneously, and there are overlaps between the phases as well. 
Also, activities across phases may be carried out simultaneously, when all of their dependent milestones were completed. Figure \ref{fig1}, shows an example Gantt chart with representative phases. Vertical lines represent the days to complete each phase, relative to the start date. Overlapping phases may be observed between for instance ``Supply and Logistics'' and ``Build'' phases or the ``Build'' and ``Integration and Acceptance'' phases. The time taken for each subsequent milestone is dependent on the time it took for previous milestones. In Figure \ref{fig1}, the time to complete a milestone in the ``Supply and Logistics'' phase depends on the completion times of its preceding milestones, occurring in the ``Solution Design'' and ``Site Acquisition'' phases. 

A project manager maintains a ``rule-of-thumb'' estimation on the time duration between a target milestone and a source milestone. Such estimates are typically additive. This domain understanding suggests that a linear (regression) model may be effective at forecasting time to completion. A network rollout requires effective planning of crew, materials, and processes for achieving the numerous milestones with minimal delays and costs. Two key data-challenges encountered are -

\label{data_chal}
\begin{enumerate}
    \item Noise in data:
    Due to differences in operating practices, project sites exhibit different standards of data reporting, integrity and level-of-detail. This results in different levels of completeness and accuracy of data across project sites. As a consequence, the approach proposed in this paper needs to deal with incomplete and noisy data; uncertainty in data needs to be treated at-least implicitly i.e., providing prediction bounds instead of point-predictions.
    \item Long-tail distribution:
    Noise in the data is not easily differentiable. There may be genuine delays induced by bad weather conditions, material shortages, or regulatory delays stemming from a bird's nest in the cell tower. Standard outlier detection methods cannot differentiate between a genuine delay and inaccurate data. Causes for delays are typically not recorded in a manner that would allow for their treatment through the modeling approach.
\end{enumerate}
Section \ref{approach} presents our proposed approach that overcomes the aforementioned domain and data challenges, while providing accurate predictions and uncertainty-bounds of duration to milestone-completion over different time-horizons.

\section{Approach}
\label{approach}

To develop a forecasting approach given attributed time-series data, the approach proposed in this paper fuses partitioning-based approaches, such as Decision Trees (DTs), ensemble models based on it - Random Forests (RFs) and Gradient Boosting (GB) and clustering/neighborhood-based approaches, with statistical approaches. Data are partitioned based on categorical attributes and regression models are developed within each data-partition.

Linear models for regression were chosen on the basis of both domain-understanding (see Section \ref{sec_prob_defn_n_challenges}) and exploratory data analysis analyzing correlations between time to complete individual milestones. In other problem contexts, depending on data characteristics and modeling needs, more complex regression models like Support Vector Machines, Neural Networks, etc. may also be used. However, in such cases, the model's explain-ability may impact its practical utilization. Also, due to generally limited training data-set sizes being available for such problem contexts and the preference for simpler, yet effective and domain-inspired models, Deep Learning methods (e.g., based on LSTMs) were not pursued within this study.

For data that conforms to a linear model, Ordinary Least Squares regression (OLS) provides a Best Linear Unbiased Estimator \cite{johnson2014applied}, as long as the underlying assumptions are not violated. The data used in this study exhibited both multicollinearity and heteroscedasticity. A delay in completion of any intermediate milestones will impact the prediction for the target milestone, and this cascading effect must be factored in the model. Hence, feature elimination was not an option. Ridge Regression (RR) was tested instead of OLS to mitigate the effects of multicollinearity. To further handle data exhibiting heteroscedasticity and provide predictive-bounds instead of point-estimates, Quantile Regression (QR) was adopted as the regression method of choice, in the proposed approach. Given the earlier discussion on data being noisy, median-based metrics were preferred (over mean-based metrics) both in the cost functions of modeling approaches and in performance metrics for their evaluation, to reduce the sensitivity of the proposed approach and its performance metrics to outlier data.

Consider that data is split into $k$ partitions, which are totally exhaustive. Let $P_i$  represent a $i^{th}$ partition, created by the partitioning algorithm and $f_\alpha^i$, a statistical function that was trained using the data points belonging to the partition $P_i$ $(\{\textbf{X}_i, \textbf{y}_i\})$ and can predict the $\alpha^{th}-$quantile value for a given $x$. The cumulative density function of the prediction/target, for a given input (test) datum, $x'$ is given by the following equation.

\begin{equation}
\label{eq_partition}
    Q_\alpha(x'| f_\alpha(\textbf{X}_i,\textbf{y}_i)) = \sum_{i=1}^k 1_{\{x'\in P_i\}} . 
    f_\alpha^i(x')
\end{equation}

Two types of data-partitioning strategies were considered in this paper. These may be understood as being ``agglomerative'' or ``discriminative'' in nature.
\begin{enumerate}
    \item Agglomerative partitioning methods identify subsets of data based on similarity in input-space. Methods used in this paper include clustering and k-nearest neighbors (k-NN). The k-NN approach is a limiting case of agglomerative partitioning in that it would create potentially overlapping partitions based on similarity in input-space. A data-structure (e.g., KD-tree) that facililates nearest-neighbor search is developed using the training data. During inference, the k-nearest training data for each test point are identified. A prediction is obtained using a Quantile Regression model learned from the identified nearest-neighbor data points.
    \item Discriminative partitioning methods strive for homogeneity relative to target attribute. The partitioning feature would minimize a prediction metric on the target (e.g., squared loss). Tree based approaches (e.g., Decision Trees) use this approach to perform data-partitioning.
\end{enumerate}
Intuitively, the agglomerative approach may produce data-partitions that are more representative of the whole dataset, capturing the spread in the target attribute, may be less prone to over-fitting, subject to the model and its hyper-parameters and may be suited for producing predictive bounds. The latter approach, in contrast, appears to be more suited for middle-point (mean/median prediction) estimation.

A summary of the models proposed and their relationships, is depicted in Table \ref{tab:models_summary}. Partitioning features refers to the type of features used for partitioning the dataset. These models are benchmarked with the ensemble models such as Random forest regression, Gradient boosting regression and Quantile Random Forest to understand whether using statistical models within partition-based regression models show competitive performance or not.

\begin{table}[h]
  \centering
  \caption{Models summary. Disc and Aggl refers to the Discriminative and Agglomerative partitioning approach.  categorical features are denoted as CAT.}
    \begin{tabular}{| >{\centering\arraybackslash}m{5em} | >{\centering\arraybackslash}m{4em} | >{\centering\arraybackslash}m{5em} | >{\centering\arraybackslash}m{4em} | >{\centering\arraybackslash}m{5em} | }
    \hline
    Models & Partitioning type & Partitioning algorithm & Partitioning features &  Base estimator\\
    \hline
    Quantile Tree & DISC & Decision Tree (CART)  &  All & QR \\
    \hline
    Piecewise QR & AGGL & Clustering (K-Means)  & CAT  & QR \\
    \hline
    Piecewise RR & AGGL & Clustering (K-Means)  & CAT  & RR\\
    \hline
    Nearest Neighbor QR & AGGL & Nearest Neighbor (KD-tree) & CAT  & QR\\
    \hline
    \end{tabular}%
  \label{tab:models_summary}%
\end{table}%

\section{Related work}
The earliest work in the decision tree for regression tasks was addressed by \cite{morgan1963problems}. The idea was to keep partitioning the feature space while there was a reduction in the loss (a parent node's loss metric was greater than sum of the loss metric of its children). The loss function is sum of the squared deviations from the mean.

The first Piecewise-linear Decision Tree (DT) algorithm - M5 \cite{quinlan1992learning} was proposed by Quinlan. Initially, the algorithm grows a sufficiently deep (standard) unpruned DT. Then, Linear Regression models were fit at each interior node using the data attributes, which were used as splitting features in the subtree below the respective node. Subsequently, a backward stepwise regression methodology was used to reduce the number of predictors of the linear models. The subtrees below the interior nodes, which showed reduced expected error based on linear models were pruned. During prediction, the leaf node's linear model output was subject to weighted averaging with the nodes along the path back to the root node. Wang and Witten extended Quinlan’s M5 model tree and named it M5$'$  \cite{frank1998using}.
In this extended version, the handling of missing values and categorical attributes were integrated on top of the M5 algorithm.

In \cite{loh2002regression}, Loh studied different methods for splitting condition within the nodes in a tree.
The paper demonstrated that it would be beneficial to fit a linear model instead of a single feature split when input features were correlated with target variable. The GUIDE algorithm, proposed in that paper, used a Chi-square test to measure the significance of each predictor or a pair of them, on the target variable. Using the chosen predictor, a linear model was fit. Data points were then classified as having either positive or negative residuals. A contingency table was created with residuals as rows and categories as columns for each attribute. The splitting variables among the predictors were chosen using the Chi-square test.
The numerical variables were split into four categories based on quantiles. A detailed comparison of various regression-tree algorithms was studied in \cite{loh2014fifty}. The approach proposed in this paper is similar to the partitioned models using Linear Regression such as M5, M5', and GUIDE. However, our approach uses models and metrics designed to cope with noisy time-series data, combining categorical attributes and numeric features.

A non-parametric approach to estimating the entire conditional distribution of the target variable was described by \cite{meinshausen2006quantile}. Our approach differs from Quantile Regression Forest (QRF) \cite{meinshausen2006quantile} in the method of calculating quantiles, as shown by Equation \ref{eq_partition}. Our approach uses Quantile Regression models inside each leaf node, whereas QRF computed the quantiles empirically, across the trees. 

Recently, a method to estimate incident duration using hybrid Decision Trees based on Quantile Regression \cite{koenker2001quantile} was proposed in \cite{he2013incident}. The GUIDE algorithm (Decision Trees) based on incident characteristics was used for partitioning of the input space. Then, a Quantile Regression model was used in each leaf of the Decision Tree based on traffic variables as predictors.
The approach presented in \cite{he2013incident} is similar to Luo's GUIDE algorithm, but used Quantile Regression models to compute the conditional distribution of the target variable. 
The cited work \cite{he2013incident} and the models discussed in our paper share a similarity in using the Quantile Regression model inside each partition, but differ in the way partitions are computed.
Another differentiating aspect is that Decision Tree based on the Classification and Regression Trees (CART) algorithm \cite{breiman1984classification} was used in our models, whereas the GUIDE algorithm was used for partitioning in \cite{he2013incident}. In  \cite{guryanov2021efficient}, an efficient algorithm for calculating SHAP values for Piecewise linear Decision Trees and ensemble models based on this, was proposed. The cited paper proved that time complexity of an algorithm to compute SHAP values for Piecewise linear DTs would be bounded by $O(MLD^2)$, where $M$, $L$ and $D$ represent the number of features, number of leaf nodes and depth of the tree.

\section{Experiments and Results}
\label{experiments}

\subsection{Data description}

\begin{figure}[h]
\centering
\includegraphics[width=0.47\textwidth]{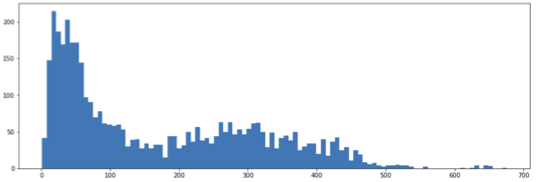}
\caption{A histogram on number of days to complete build-phase-milestone-1 from build-phase-milestone-2. Major portion of the project's had the duration within 100 days.}
\label{target_dist}
\end{figure}

The data comprised of the status of different projects being executed across several sites. Each site was represented by categorical attributes such as city, state, region, market categories, and latitude and longitude information.
Few sites had one or more projects running at the same time.
Similarly, metadata information for projects included nature of the project, technology being deployed, and a list of phases involved in that particular project. 
The data collected on completed projects include start and end dates for every milestone in the respective projects. There were 2066 projects which contained both the source and target milestones of interest. Each project had target (planned) and actual completion times for the achieved milestones, depending on the extent of completion of the project. Figure \ref{target_dist} shows a histogram of the unprocessed time difference between the completion time of two milestones in the Build phase. The data exhibited a long-tail (right-skewed) distribution, showcasing the challenges mentioned in the Section \ref{sec_prob_defn_n_challenges}. Data were pre-processed prior to modeling and evaluation - details are provided in Appendices \ref{sec_feature_engg}, \ref{sec_data_wrangling} and \ref{sec_feature_selection}.

\subsection{Models evaluated}

For performance evaluation, ten models were considered, chosen in accordance with section \ref{approach}. The approaches defined are described in the Table \ref{tab:models_summary}. The best hyper-parameters for each models were chosen based on a five-fold cross-validation on the entire dataset. The models were fit with the best performing hyper-parameter combination on the entire training dataset. The hyper-parameters that were tuned across models include:
\begin{enumerate}
\item Regularization penalty values for Ridge Regression, Quantile Regression (L1-penalty), Quantile Trees, piecewise models, and Nearest Neighbor Quantile Regression
\item Minimum number of samples to split and maximum depth of the tree for Decision Tree and Random Forest Regression
\item Number of estimators and learning rate for Gradient Boosting models
\item Number of clusters for piecewise linear models
\end{enumerate}

\subsection{Evaluation and results}

The experiments performed sought to evaluate the performance of the proposed partitioned regression models and compare them with statistical models or tree-based ensemble models. In this paper, we show three experiments over significantly different time-frames. The first experiment has start and end milestones within the same phase (specifically, the Build phase), but the duration between the two milestones ranged up-to 120 days. The second experiment contained the start and end milestones from different phases - time duration from a site-acquisition-phase milestone to build-phase milestone. This experiment was considered for long-range forecasting (up to 350 days). For the third experiment, two milestones were chosen from the same phase (Build), which were performed relatively close to each-other (around 60 days). This experiment tested the proposed approach in the context of short-range forecasting.

\begin{table}[h!]
  \centering
  \caption{Model performance for predicting a build-phase target milestone from a build-phase source milestone, given time duration between them can be up to 120 days. Quantile Tree model achieved almost same performance as Gradient Boosting Regressor while having 7\% only in number of the parameters}
    \begin{tabular}{lccc}
    \hline
    \makecell{Approach}& \makecell{Median AE \\ (in days)} & \makecell{Mean AE \\ (in days)}  & \makecell{Number of\\ parameters}\\
    \hline
    Ridge Regressor  & 6.79  & 9.55 & 51\\
    Quantile Regressor & 5.43 & 9.18  & 51\\
    Decision Tree Regressor & 7.4  & 11.12 & 20 \\
    Random Forest Regressor & 8.19 & 11.62   & 2000\\
    QRF & 8.14  & 11.71& 2000\\
    \textbf{Gradient Boosting Regressor}  & \textbf{4.85} & \textbf{8.35}  & 2000\\
    \textbf{Quantile Tree}  & \textbf{5} & \textbf{9.69}  & 140\\
    Piecewise QR & 5.23 & 8.97  & 400\\
    Piecewise RR & 7.08 & 9.44   & 400\\
    Nearest Neighbor QR & 5.07  & 9.05  & NA\\
    \hline
    \end{tabular}%
  \label{tab:medium_range_forecasting}%
\end{table}%

\begin{table}[h!]
  \centering
  \caption{Model performance for predicting a build-phase milestone from a site-acquisition-phase  source milestone, given time duration between them can be up to 350 days. Quantile Tree model achieved the best performance while having 8\% of the model complexity of the Gradient Boosting Regressor.}
    \begin{tabular}{lccc}
    \hline
    \makecell{Approach}& \makecell{Median AE \\ (in days)} & \makecell{Mean AE \\ (in days)}  & \makecell{Number of\\ parameters}\\
    \hline
    Ridge Regressor & 10.04 & 14.47  & 52\\
    Quantile Regressor & 5.81 & 11.9 & 52  \\
    Decision Tree Regressor & 9.3 & 14.63 & 20 \\    
    Random Forest Regressor  & 9.29 & 14.57 & 2000\\
    QRF & 9.04 & 14.52  & 2000\\
    \textbf{Gradient Boosting Regressor}  & \textbf{4.89}  & \textbf{9.13} & \textbf{2000}\\
    \textbf{Quantile Tree}  & \textbf{0} & \textbf{5.5}   & \textbf{160}\\
    Piecewise QR  & 5.26 & 11.74 & 408\\
    Piecewise RR & 9.62 & 14.28  & 408\\
    Nearest Neighbor QR  & 5.71 & 11.51 & NA\\
    \hline
    \end{tabular}%
  \label{tab:long_range_forecasting}%
\end{table}%

\begin{table}[h!]
  \centering
  \caption{Model performance for predicting a build-phase target milestone from a build-phase source milestone, given time duration between them can be up to 60 days. Quantile tree achieved the best performance while having 4\% of the model complexity of the Gradient Boosting Regressor}
    \begin{tabular}{lccc}
    \hline
    \makecell{Approach}& \makecell{Median AE \\ (in days)} & \makecell{Mean AE \\ (in days)}  & \makecell{Number of\\ parameters}\\
    \hline
    Ridge Regressor & 5.52 & 7.02  & 48 \\
    Quantile Regressor  & 5.02 & 7.34 & 48 \\
    Decision Tree Regressor  & 5.2 & 7.54 & 20 \\
    Random Forest Regressor  & 5.38 & 7.72 & 2000 \\
    QRF  & 5.46  & 7.75 & 2000 \\
    \textbf{Gradient Boosting Regressor }  & \textbf{4.43} & \textbf{6.56}  & \textbf{2000} \\
    \textbf{Quantile Tree} & \textbf{4} & \textbf{6.72}  & \textbf{80} \\
    Piecewise QR & 5  & 7.3  & 376 \\
    Piecewise RR & 6.14  & 7.71 & 376 \\
    Nearest Neighbor QR & 4.3 & 6.9  & NA  \\
    \hline
    \end{tabular}%
  \label{tab:short_range_forecasting}%
\end{table}%

Tables \ref{tab:medium_range_forecasting}, \ref{tab:long_range_forecasting}, and \ref{tab:short_range_forecasting} summarize the five-fold cross-validation performance of the models benchmarked against each-other after hyper-parameter tuning, on both mean and median absolute error metrics. The median metric is our primary evaluation metric due to the noise in the data but mean was recorded for completeness.

Simple statistical models such as Ridge Regression and Quantile Regression provided better performance than the Decision Tree and Random Forest regression models. This could be because statistical models were more effective at modeling the linear relationship between the predictors and the target variable. For instance, it can easily incorporate the aspect of cascading effect of delays in intermediate milestones, unlike Decision Trees and Random Forests.
Quantile Tree (QT) uses a one single decision tree of depth eight, whereas the tree ensemble models utilize 100 trees, each of depth 4-5, to achieve competitive performance. The drastic difference in the model complexity between the QT and Gradient Boosting regressor is evident from the number of parameters trained (140 to 2000) as part their modelling.
From all three experiments, Quantile Tree was observed as one of the top three performing models, while having among the least model complexities.
The table \ref{tab:long_range_forecasting} shows that Quantile Tree achieved top performance in the long range forecasting experiment with zero median absolute error using single Decision Tree enhanced with Quantile Regression models, as described in this paper.

In short-range forecasting, where the number of intermediate milestones were fewer, all the models did comparably well. The advantage of incorporating linear models within partitioned regressors was reduced, due to the lesser number of numerical variables. In long-range forecasts, where the number of intermediate milestones was higher, linear models within partitioned regressors performed better than ensemble models. The proposed models were able to provide competitive (within $\approx$0.5 day median error over a 60 or 120 day prediction range, relative to the next best approach) to better (a $\approx$5 day reduction in median error over a 350 day prediction range, compared to the next best approach) prediction accuracy for both long-range and short-range forecasts, at a small fraction of the model complexity of the next best approach, in each case. The forecast error was within four to five days because linear models were more effective due to more intermediate milestones.

\begin{figure}[htp]
\centering
\includegraphics[width=0.48\textwidth]{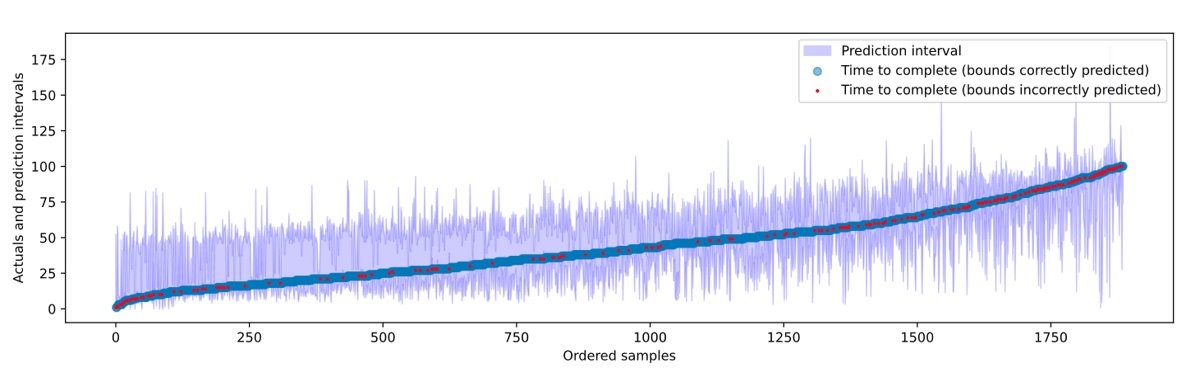}
\caption{Predicted bounds and actual values of time to complete the target milestone. The Blue and Red dots represent the actual values, which are within and outside the predicted bounds respectively.}
\label{pred_bounds}
\end{figure}

To obtain confidence bounds, the $5^{th}$ and $95^{th}$ percentiles were used in QR to compute the lower and upper bounds. 
After computing the bounds, the true values of the target attribute of interest (time to complete the target milestone) were compared against the predicted bounds. The model achieved 88.72\% accuracy in providing a bound that contained the true value. Figure \ref{pred_bounds} showcases the predicted bounds and the actual values for number of days to complete the target milestone from the reference milestone. 


\subsection{Experiments on a Public Workloads dataset}

To demonstrate the applicability of the proposed approaches, an experiment was carried out on a public dataset from Grid Workloads Archive (\footnote{http://gwa.ewi.tudelft.nl}). The dataset contains Key Performance Indicators (KPIs) of Virtual Machines (VMs) from a distribution data-center of Materna (\cite{kohne2014federatedcloudsim}, \cite{kohne2016evaluation}) at a five minutes interval. 
Forecasting these KPIs is one of the important problem, which is essential in the Tele-communications domain. Based on this Virtual Machines workloads dataset, CPU usage [MHZ] was forecasted using a list of predictor features. The predictors were number of CPU cores, CPU capacity provisioned [MHZ], Memory capacity provisioned [KB], CPU usage [MHZ], Disk size [GB], Memory usage [KB], Disk read throughput [KB/s], Disk write throughput [KB/s], Network transmitted throughput [KB/s] and Network received throughput [KB/s]. The hour of the day and day of the week were computed from the date timestamps and these features were used as categorical attributes along with number of CPU cores. Three lags of the target variable were also used as the predictors (number of lags was decided on a three fold-cross validation scoring on Quantile Regression and Linear regression model) to accommodate the auto-regressive component of the modelling.

The model performances were calculated from first twenty VMs for this experiment, which accounts for 1.5M datapoints. 
After hyper-parameter tuning for respective models, the best performance achieved are reported for the each model in the Table \ref{tab:public-dataset}. The results showcase that the QT model is just 0.09 MHz in Median AE (1.36 MHz in Mean AE) behind the Gradient Boosting regressor model, at the same time requiring only one decision tree to achieve such competitive performance. The Piecewise models were also not far behind, especially Piecewise QR model was able to attain the third best performance, which explains the advantage of combining the partitioning and QR models.

\begin{table}[h!]
  \centering
  \caption{Model performance on CPU Usage prediction for VMs}
    \begin{tabular}{lcc}
    \hline
    \makecell{Approach}& \makecell{Mean AE \\ (in MHZ)} & \makecell{Median AE \\ (in MHZ)} \\
    \hline
    Ridge Regressor & 30.19 &  7.35 \\
    Quantile Regressor  & 27.83 & 4.60 \\
    Decision Tree Regressor  & 22.44 & 3.29 \\
    Random Forest Regressor  & 40.13 & 4.17\\
    \textbf{Gradient Boosting Regressor }  & \textbf{21.09} & \textbf{3.21}\\
    \textbf{Quantile Tree} & \textbf{22.45}  & \textbf{3.30} \\
    Piecewise QR & 25.97  & 4.28 \\
    Piecewise RR & 29.26  & 7.64\\
    Nearest Neighbor QR & 23.07 & 5\\
    \hline
    \end{tabular}%
  \label{tab:public-dataset}%
\end{table}%

\section{Conclusion}

This paper presented a data and ML-driven approach to predict time-to-complete milestones for the Telecommunications network-rollout planning problem. The problem involved modeling attributed time-series data, and the paper proposed statistical models within partitioned regression methods as an effective solution to the forecasting problem. The outcomes of this paper demonstrate that statistical models within data partitioning-based regression methods effectively captures data trends, tends to be at-least as performant as the best-known approaches, at a small fraction of their model complexity. Leveraging data-trends can enable the simplest models to be competitive or better than sophisticated general-purpose approaches. A solution deployed based on the proposed approach provided compelling prediction errors for both short-range and long-range forecasts, suggesting that the proposed approach would be viable for enabling subsequent network-rollout planning.

\bibliographystyle{IEEEtran}
\bibliography{refs}

\appendix

\section{Feature Engineering}
\label{sec_feature_engg}
Based on inputs from domain experts, the following additional features were computed from existing data attributes.
These features provided useful information for the model to predict the time required to complete the target milestone.

\begin{enumerate}
    \item Derived date features, such as a month, quarter, and year, were computed from the source milestone's completion date.
    \item A feature that encoded a site's climate classification was incorporated. Derived features from the date and climate, were together used as a surrogate for the temperature or local weather conditions.
    \item To represent regional regulatory trends, the first two digits of a site-location's zip or postal code was extracted to represent the broad geographical region of the site.
    \item  A set of derived features were computed to represent the time to complete intermediate milestone, relative to a start milestone. These numeric features were long-tailed due to both inaccuracies in the data and genuine cases of delays. Based on SME feedback on meaningful upper-limits, these long-tails were pruned. 
\end{enumerate}

\section{Data Wrangling}
\label{sec_data_wrangling}
The following data-processing steps were performed to handle the challenges explained in the Section \ref{data_chal}. Inferences from this analysis on sequential dependencies between milestones were vetted by domain experts before being used for modeling.
\begin{enumerate}
    \item Ranking analysis was done on the milestones within the phase to identify the typical ordering of the milestones in each phase.
    \item A matrix was created to represent the aggregated time difference (mean and median) between each pair of activities across different projects. This matrix helped measure the expected time-gap between two milestones. This was used to identify a potential list of intermediate milestones. 
    \item Correlation analysis between the time-to-complete earlier milestones and the target milestone was done. A delay in completing an intermediate milestone has a cascading effect on successive milestones, including the target. \textbf{This signified the existence of multicollinearity between the numerical features in data}. A list of intermediate milestones was chosen based on the correlation analysis in such a way that all the intermediate milestones would be completed before the start of the target activity. 
    \item Outlier projects were removed on the basis of the completion times of both the target and intermediate milestones. This was done to filter out obvious errors due to inaccurate data entered into the system.
\end{enumerate}

\section{Feature Selection}
\label{sec_feature_selection}

Feature selection was performed on more than a hundred features (including both categorical and numerical features) about the project and site meta-data.
To perform feature selection, the Chi-square test \cite{loh2002regression} was used for categorical attributes and Pearson's correlation coefficient for numerical features.
After feature selection, there were eleven categorical attributes, including the derived features from the starting milestone date and climate classification of the site's state. The number of numerical features varied between four and ten, based on the range (short or long term) of forecasting. This reduced feature set aids interpretability of the forecasting model.

\end{document}